\pdfoutput=1

\documentclass[11pt]{article}

\usepackage[final]{acl}
\usepackage{tcolorbox}
\usepackage{amsmath}
\usepackage{booktabs}
\usepackage{multirow}
\usepackage{subcaption}
\usepackage{xcolor}  
\usepackage{times}
\usepackage{latexsym}
\usepackage{amsfonts}
\usepackage{kotex}
\usepackage{colortbl}
\usepackage[T1]{fontenc}

\usepackage[utf8]{inputenc}

\usepackage{microtype}

\usepackage{inconsolata}

\usepackage{color}
\usepackage{tabularray}

\usepackage{graphicx}

\usepackage{enumitem}

\usepackage{CJKutf8}

\usepackage{xspace}

\usepackage{siunitx}
\usepackage{threeparttable}
\title{\textsc{LangSAE Editing}: Improving Multilingual\\Information Retrieval via Post-hoc Language Identity Removal}

\author{
  \textbf{Dongjun Kim}$^{1}$\thanks{Equal contribution.}\ ,
  Jeongho Yoon$^{1}$\footnotemark[1]\ ,
  Chanjun Park$^{2}$\thanks{Corresponding author.}\ ,
  Heuiseok Lim$^{1}$\footnotemark[2] \\
  $^{1}$Department of Computer Science and Engineering, Korea University \\
  $^{2}$School of Software, Soongsil University \\
  \texttt{\{junkim100, aa007878, limhseok\}@korea.ac.kr} \\
  \texttt{chanjun.park@ssu.ac.kr}
}

\begin{document}
\maketitle

\begin{abstract}
Dense retrieval in multilingual settings often searches over mixed-language collections, yet multilingual embeddings encode language identity alongside semantics. This language signal can inflate similarity for same-language pairs and crowd out relevant evidence written in other languages. We propose \textsc{LangSAE Editing}, a post-hoc sparse autoencoder trained on pooled embeddings that enables controllable removal of language-identity signal directly in vector space. The method identifies language-associated latent units using cross-language activation statistics, suppresses these units at inference time, and reconstructs embeddings in the original dimensionality, making it compatible with existing vector databases without retraining the base encoder or re-encoding raw text. Experiments across multiple languages show consistent improvements in ranking quality and cross-language coverage, with especially strong gains for script-distinct languages. The \textsc{LangSAE} model and training code are publicly available.\footnote{https://github.com/junkim100/LangSAE-Editing}
\end{abstract}

\section{Introduction}

Dense retrieval ranks documents by comparing query and document embeddings, typically with cosine similarity, and it is a core component of modern search and retrieval-augmented generation pipelines \cite{karpukhin-etal-2020-dense,xiong-etal-2021-ance,khattab-zaharia-2020-colbert}. In multilingual deployments, the indexed collection is often mixed-language, and relevant evidence for a query can appear in any language. In this setting, dense retrievers commonly exhibit a \textit{same-language preference}, where same-language candidates receive a similarity advantage and crowd out more relevant evidence written in other languages \cite{yang2021simple,yang-etal-2025-language-bias-ir}.

We study this setting as multilingual information retrieval (MLIR): queries may be issued in any supported language and retrieval is performed against a single multilingual pool \cite{zhang-etal-2023-miracl,zhang-etal-2021-mr}. The failure mode is a mismatch with the goal of embedding-based retrieval, which is to prioritize semantic alignment rather than language match.

Prior analyses point to a concrete mechanism. Multilingual encoders encode language identity in addition to semantics, and language identity remains recoverable from their representations \cite{devlin-etal-2019-bert,conneau-etal-2020-unsupervised,libovicky-etal-2020-language,libovicky-etal-2019-how}. When similarity search is performed in a shared space, this language signal can distort neighborhood structure by inflating same-language cosine similarity, producing \textbf{Language Identity Bias in MLIR} \cite{yang2021simple,yang-etal-2025-language-bias-ir}.

Mitigating this bias is constrained by deployment reality. Many systems rely on precomputed document embeddings in a vector database, and encoder-side mitigation typically requires fine-tuning and then re-encoding the entire corpus from raw text, which is often the dominant cost in real deployments. This motivates post-hoc methods that operate directly on existing vectors and remain compatible with standard similarity search infrastructure \cite{johnson-etal-2019-faiss,malkov-yashunin-2020-hnsw}.

We propose \textsc{LangSAE}, a post-hoc method that suppresses language-identity signal in pooled embeddings while preserving retrieval-relevant semantics. \textsc{LangSAE} is an overcomplete sparse autoencoder trained on pooled embeddings, its sparse feature representation enables language-associated factors to concentrate into a small set of latent units that can be selectively suppressed, then decoded back to the original embedding dimensionality for drop-in cosine scoring. Because the transformation is vector-only, it is substantially cheaper than encoder tuning and corpus-wide re-encoding, editing an embedding in 0.0445\,ms, enabling both offline retrofitting of stored vectors and query-time editing.

Across Belebele \cite{bandarkar-etal-2024-belebele} and XQuAD \cite{artetxe-etal-2020-cross}, \textsc{LangSAE Editing} improves macro-average nDCG@20 by about +21.9\% and +20.6\%, respectively. Gains are especially large for script-distinct languages such as Chinese, consistent with language identity acting as a similarity shortcut in multilingual pools.

We make three contributions:
\begin{itemize}
    \item We formalize \textbf{Language Identity Bias in MLIR} as same-language crowding in shared multilingual pools and introduce diagnostics that isolate this effect beyond aggregate retrieval metrics.
    \item We introduce \textsc{LangSAE Editing}, a sparse feature-based post-hoc transformation that suppresses language-associated units and reconstructs embeddings in the original space for drop-in retrieval.
    \item We demonstrate consistent gains on multilingual pools across two benchmarks and provide analyses that connect feature suppression to improved ranking behavior, with a lightweight transformation that is practical for retrofitting existing vector databases.
\end{itemize}

\section{Related Work}

\subsection{Multilingual Dense Retrieval}
Dense retrieval embeds queries and documents into a shared space and ranks by vector similarity \citep{karpukhin-etal-2020-dense}. Multilingual retrievers typically build on pretrained multilingual encoders \citep{devlin-etal-2019-bert,conneau-etal-2020-unsupervised} and are evaluated on multilingual retrieval datasets such as mMARCO, Mr.~TyDi, and MIRACL \citep{bonifacio-etal-2021-mmarco,zhang-etal-2021-mr,zhang-etal-2023-miracl}, with broad embedding evaluations increasingly standardized by MTEB \citep{muennighoff-etal-2022-mteb}. Recent improvements come from multilingual sentence embedding alignment \citep{artetxe-schwenk-2019-massively,feng-etal-2022-language}, weakly supervised contrastive pretraining \citep{wang-etal-2022-e5,wang-etal-2024-multilingual-e5}, unsupervised pretraining for multilingual dense retrieval \citep{wu-etal-2022-unsupervised}, lightweight inference-time adaptation \citep{huang-etal-2023-soft-prompt}, contrastive objectives for language-agnostic retrieval \citep{hu-etal-2023-language}, and distillation-based transfer \citep{yang-etal-2024-distillation}. Despite these advances, retrieval quality often varies substantially across languages, motivating analyses of language-linked failure modes in multilingual IR \citep{yang-etal-2025-language-bias-ir}.

\begin{figure*}[t]
    \centering
    \includegraphics[width=\linewidth]{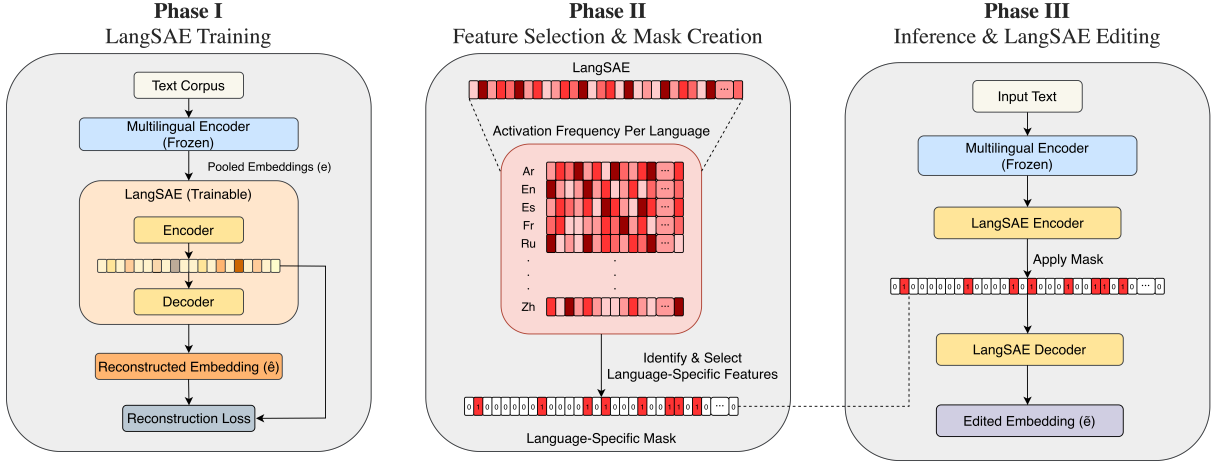}
    \caption{Overview of the \textsc{LangSAE Editing} pipeline.
    \textbf{Phase I:} Train an overcomplete sparse autoencoder on pooled embeddings from a frozen multilingual encoder.
    \textbf{Phase II:} Compute per-language activation frequencies and select language-associated features to form a mask.
    \textbf{Phase III:} Encode text, apply the mask in latent space, and decode to obtain an edited embedding for retrieval.}
    \label{fig:method_pipeline}
\end{figure*}

\subsection{Language Signal and Bias in Multilingual Representations}
Language identity remains recoverable from multilingual representations, indicating that embeddings mix semantics with language-correlated structure \citep{libovicky-etal-2019-how,libovicky-etal-2020-language}. Related work studies when cross-lingual transfer emerges and how representation spaces align across languages \citep{artetxe-etal-2020-cross,artetxe-schwenk-2019-massively}, and proposes reducing self-language preference by explicitly removing language information \citep{yang2021simple}. Other approaches encourage language-agnostic structure during training \citep{zhao-etal-2021-inducing} or identify language-associated subspaces that can be filtered \citep{xie-etal-2022-discovering}, while recent dense retrieval work explores language-invariant behavior through language concept erasure \citep{huang-etal-2024-language}. From an IR perspective, language bias is also framed as an evaluation and fairness issue, where per-language reporting and disparity-aware analysis are important \citep{bandarkar-etal-2024-belebele,yang-etal-2025-language-bias-ir}.

\subsection{Post-hoc Representation Transformation}
Post-processing methods can improve cosine-neighborhood geometry by addressing anisotropy or dominant directions in embedding spaces, often via simple transformations such as whitening \citep{li-etal-2020-sentence,huang-etal-2021-whiteningbert-easy}. Autoencoder-based objectives offer an alternative, learning a transformation that reconstructs vectors while enabling controlled edits in latent space, as demonstrated by reconstruction-based sentence embedding learning \citep{wang-etal-2021-tsdae-using}. Our work follows this post-hoc direction but targets a specific nuisance factor, language identity, via sparse overcomplete features that can be selectively suppressed while reconstructing embeddings back to the original dimensionality for drop-in retrieval use.

\section{Methodology}
\label{sec:method}

We propose \textsc{LangSAE}, a post-hoc method that edits pooled encoder embeddings to suppress language-identity signal while preserving retrieval-relevant semantics. We assume a standard dense retrieval pipeline where a frozen multilingual encoder produces token representations, which are mean-pooled into a single vector. Queries and documents are ranked by cosine similarity between $\ell_2$-normalized pooled embeddings \citep{karpukhin-etal-2020-dense}. In multilingual information retrieval (MLIR), the candidate pool mixes multiple languages, and language identity encoded in embeddings can inflate similarity for same-language pairs, leading to same-language crowding in the top ranks. \textsc{LangSAE Editing} mitigates this effect by first rewriting pooled embeddings into an overcomplete \emph{sparse} feature representation, where each embedding is expressed by a small set of active latent units. This representation makes language-related signal identifiable as consistently activated units within each language, enabling selective suppression without applying a single global transformation uniformly to all inputs. The edited representation is then decoded back to the original embedding dimensionality, so retrieval remains a drop-in replacement for existing cosine similarity search and vector database infrastructure \citep{li-etal-2020-sentence,huang-etal-2021-whiteningbert-easy} (Figure~\ref{fig:method_pipeline}). Because the transformation operates purely on vectors, it can be applied both to retrofit stored document embeddings offline and to transform query embeddings at runtime, without modifying the underlying encoder or requiring access to raw text.

\subsection{Phase I: Training \textsc{LangSAE} on pooled embeddings}
\label{sec:langsae_training}

Let $\mathbf{e}\in\mathbb{R}^d$ denote the \emph{raw} pooled embedding produced by a frozen base encoder for a text segment. \textsc{LangSAE} is trained directly on raw pooled embeddings. At retrieval time, reconstructed and edited embeddings are $\ell_2$-normalized before cosine similarity scoring (Section~\ref{sec:exp_setup}).

\textsc{LangSAE} is an overcomplete sparse autoencoder with encoder $E_\theta$ and decoder $D_\phi$, where the latent dimensionality is $m \gg d$; sparsity encourages reusable latent units and makes activation-frequency statistics meaningful for isolating language-associated features. Given $\mathbf{e}$, the encoder produces latent pre-activations, followed by a ReLU nonlinearity to obtain non-negative activations:
\begin{equation}
\mathbf{z} = \mathrm{ReLU}(E_\theta(\mathbf{e})) \in \mathbb{R}^{m}, \qquad z_i \ge 0 \ \ \forall i.
\label{eq:relu_latent}
\end{equation}
We impose sparsity by keeping only the top-$k$ activated features per example (top-$k$ applied directly to ReLU activations) \citep{makhzani-frey-2013-k-sparse}. Let $\mathrm{TopK}(\mathbf{z},k)$ return the indices of the $k$ largest entries of $\mathbf{z}$. The sparsification operator $S(\cdot)$ is:
\begin{equation}
\left[S(\mathbf{z})\right]_i =
\begin{cases}
z_i, & i \in \mathrm{TopK}(\mathbf{z},k), \\
0, & \text{otherwise.}
\end{cases}
\label{eq:topk_sparsify}
\end{equation}
We denote the resulting sparse code by
\begin{equation}
\tilde{\mathbf{z}} = S(\mathbf{z}) = S(\mathrm{ReLU}(E_\theta(\mathbf{e}))).
\label{eq:sparse_code_def}
\end{equation}

\textsc{LangSAE} is trained to reconstruct pooled embeddings with mean squared error:
\begin{equation}
\mathcal{L}_{\mathrm{rec}}(\theta,\phi)
= \mathbb{E}_{\mathbf{e}}
\left[
\|\mathbf{e} - D_\phi(\tilde{\mathbf{z}})\|_2^2
\right].
\label{eq:recon_loss}
\end{equation}
We additionally include sparsity-related auxiliary terms to encourage stable sparse features.

\subsection{Phase II: Identifying language-associated latent features}
\label{sec:lang_feature_identification}

After training, we identify language-associated latent units using activation statistics computed on a language-labeled probe set. Let $\mathcal{P}_\ell$ be probe texts in language $\ell$. For each $t\in\mathcal{P}_\ell$, we compute its pooled embedding $\mathbf{e}(t)$ and sparse code $\tilde{\mathbf{z}}(t)$ (Eq.~\ref{eq:sparse_code_def}). We treat a latent unit as \emph{active} if its sparse activation is non-zero (after top-$k$ sparsification), and estimate how often each unit activates in each language:
\begin{equation}
p_{i,\ell} \;=\; \mathbb{E}_{t\sim\mathcal{P}_\ell}\!\left[\mathbb{I}\!\left(\tilde z_i(t) > 0\right)\right].
\label{eq:activation_frequency}
\end{equation}

Given a global threshold $\tau\in(0,1]$, we construct three sets of units:
\begin{itemize}
    \item \textbf{Frequent for language $\ell$:} $\mathcal{F}_\ell(\tau)=\{i : p_{i,\ell}\ge\tau\}$.
    \item \textbf{Language-unique for $\ell$:} $\mathcal{U}_\ell(\tau)=\{i\in\mathcal{F}_\ell(\tau) : \max_{\ell'\neq\ell} p_{i,\ell'} < \tau\}$.
    \item \textbf{Overlapping:} $\mathcal{O}(\tau)=\{i : i\in\mathcal{F}_\ell(\tau)\cap\mathcal{F}_{\ell'}(\tau)\ \text{for some }\ell\neq\ell'\}$.
\end{itemize}
Intuitively, $\mathcal{U}_\ell(\tau)$ contains units that fire reliably for language $\ell$ but not for other languages, while $\mathcal{O}(\tau)$ contains units that are frequent across multiple languages, which may reflect shared scripts, tokenization regularities, or multilingual corpus artifacts. We keep these sets distinct to support different masking strategies at inference (Section~\ref{sec:langsae_editing}).

The threshold $\tau$ controls how conservatively units are selected. Values near $\tau\approx 1.0$ retain only the most consistently activated units, while lowering $\tau$ can rapidly increase the frequent sets and risk including broadly used units. We compute $p_{i,\ell}$ on the held-out validation split used during \textsc{LangSAE} training (Appendix~\ref{sec:appendix_probe_set}) and report a sensitivity sweep over $\tau$ in Appendix~\ref{sec:ablation_tau}. We also evaluate whether including overlapping units in the suppression set is beneficial (Appendix~\ref{sec:ablation_overlap}).

\subsection{Phase III: \textsc{LangSAE Editing} at inference}
\label{sec:langsae_editing}

\textsc{LangSAE Editing} requires a language label $\ell$ for each embedding. In our benchmarks, $\ell$ is provided by the dataset. In deployed systems, $\ell$ can come from document metadata or a standard language identification module.

Given a pooled embedding $\mathbf{e}$ and its language $\ell$, we first compute its sparse latent code $\tilde{\mathbf{z}}$ using the trained \textsc{LangSAE} encoder and the top-$k$ sparsification defined in Section~\ref{sec:langsae_training}. We then remove language-associated signal by masking a selected set of latent units derived from the activation-frequency statistics in Section~\ref{sec:lang_feature_identification}. We consider two masking strategies:
(i) \emph{Unique-only}, masking $\mathcal{U}_\ell(\tau)$, and
(ii) \emph{Unique+Overlapping}, masking $\mathcal{U}_\ell(\tau)\cup\mathcal{O}(\tau)$.
We use \emph{Unique+Overlapping} in our main experiments (with $\tau=0.999$) and report \emph{Unique-only} as an ablation (Appendix~\ref{sec:ablation_overlap}).

Concretely, \textsc{LangSAE Editing} applies the following steps:
\begin{enumerate}
    \item \textbf{Encode to sparse features:} compute $\tilde{\mathbf{z}}$ from $\mathbf{e}$ using the trained encoder and top-$k$ sparsification.
    \item \textbf{Language-conditioned masking:} form a mask set $\mathcal{S}_\ell$ (either $\mathcal{U}_\ell(\tau)$ or $\mathcal{U}_\ell(\tau)\cup\mathcal{O}(\tau)$), then set the corresponding latent coordinates to zero to obtain a masked code $\tilde{\mathbf{z}}'$.
    \item \textbf{Decode back to the original space:} reconstruct an edited embedding $\tilde{\mathbf{e}} = D_\phi(\tilde{\mathbf{z}}')$.
    \item \textbf{Normalize for cosine scoring:} output $\bar{\tilde{\mathbf{e}}} = \tilde{\mathbf{e}} / \|\tilde{\mathbf{e}}\|_2$.
\end{enumerate}

Masking can reduce the number of active features below $k$ for some examples. We keep this behavior (rather than refilling from lower-ranked activations) to avoid reintroducing correlated units.

For retrieval, we apply the same transformation (Baseline, SAE Reconstructed, or \textsc{LangSAE Editing}) to both queries and documents, and rank candidates by cosine similarity between the resulting $\ell_2$-normalized vectors:
\begin{equation}
s(q,d)=\left\langle \bar{\mathbf{e}}_q, \bar{\mathbf{e}}_d \right\rangle.
\label{eq:cosine_score}
\end{equation}

\subsection{Control: SAE reconstruction without masking}
\label{sec:controls}

To isolate the contribution of feature suppression from reconstruction, we define an SAE reconstruction control that passes embeddings through \textsc{LangSAE} without masking:
{\small
\begin{equation}
\bar{\hat{\mathbf{e}}}
=
\frac{D_\phi(\tilde{\mathbf{z}})}{\|D_\phi(\tilde{\mathbf{z}})\|_2},
\qquad
\tilde{\mathbf{z}} = S(\mathrm{ReLU}(E_\theta(\mathbf{e}))).
\label{eq:reconstruct_control}
\end{equation}
}
Comparing \textbf{Baseline}, \textbf{SAE Reconstructed}, and \textbf{\textsc{LangSAE Editing}} in Section~\ref{sec:experiments} separates the effect of sparse autoencoding from the effect of targeted language-feature suppression.

\section{Experiments}
\label{sec:experiments}

\subsection{Experimental Setup}
\label{sec:exp_setup}

\paragraph{Task.}
We evaluate mixed-language multilingual information retrieval (MLIR): each query retrieves from a single multilingual pool that contains documents from multiple languages, and relevant evidence may appear in any language.

\paragraph{Benchmarks.}
We use Belebele \citep{bandarkar-etal-2024-belebele} and XQuAD \citep{artetxe-etal-2020-cross} on 10 languages: Arabic, Chinese, English, French, Hindi, Italian, Japanese, Portuguese, Russian, and Spanish. Both benchmarks provide parallel/aligned documents across languages (Belebele via passage IDs, XQuAD via SQuAD example IDs); each passage/context paragraph is treated as one retrieval unit. More details about benchmarks can be found in Appendix \ref{sec:appendix_benchmarks}.

\paragraph{Pools and relevance.}
For each benchmark we form a multilingual pool by taking the union of documents across the included languages, and we evaluate queries grouped by query language against the same shared pool. Belebele yields $4{,}880$ documents ($488\times10$) and $9{,}000$ queries ($900\times10$). XQuAD yields $1{,}440$ documents ($240\times6$) and $7{,}140$ queries ($1{,}190\times6$) for the available 6 languages (Arabic, Chinese, English, Hindi, Russian, Spanish). For each query $q$, the multi-relevant set $R_q$ is the aligned document set across languages, so $|R_q|=10$ on Belebele and $|R_q|=6$ on XQuAD.

\paragraph{Retrieval and systems.}
Documents are not chunked at evaluation time. We encode text with mean pooling (Section~\ref{sec:method}), $\ell_2$-normalize embeddings, and rank by cosine similarity using exact (brute-force) search over the full pool. We compare: (i) \textbf{Baseline}, frozen encoder; (ii) \textbf{SAE Reconstructed}, \textsc{LangSAE} without masking; and (iii) \textbf{\textsc{LangSAE Editing}}, masking with the \textit{Unique + Overlapping} strategy at $\tau{=}0.999$. The same transformation is applied to both queries and documents.

\paragraph{Selecting $\tau$.}
We compute activation frequencies $p_{i,\ell}$ on the held-out validation split used in \textsc{LangSAE} training data preparation (Appendix~\ref{sec:appendix_training_data}, \ref{sec:appendix_probe_set}) and choose a conservative $\tau$ to avoid over-masking; we use $\tau=0.999$ unless stated otherwise (Appendix~\ref{sec:ablation_tau}).

\paragraph{Metrics.}
We report standard Recall@20 and nDCG@20 \citep{jarvelin-kekalainen-2002-ndcg} with binary relevance. In our mixed-language setting, each query $q$ has a multi-relevant set $R_q$ consisting of all aligned passages across languages (10 for Belebele, 6 for XQuAD). A retrieved passage is counted as relevant if it belongs to $R_q$. Recall@20 is the fraction of $R_q$ retrieved in the top 20. nDCG@20 is computed with binary gains and an ideal ranking that places all relevant passages first (up to 20). We report averages per query language and a macro-average that weights each query language equally.

\subsection{Bias Evidence: Quantifying Language Bias via Ground-Truth Removal}
\label{sec:bias_evidence}

\begin{table}[t]
    \centering
    \resizebox{\linewidth}{!}{%
        \begin{tabular}{lrr|r}
        \toprule
         & \multicolumn{2}{c|}{\textbf{Avg. Count @ Top-20}} & \\
        \textbf{Retrieved Lang.} & \textbf{m-e5-large} & \textbf{LangSAE} & \textbf{$\Delta$ Count} \\ 
        \midrule
        \multicolumn{4}{l}{\textit{\textbf{Query Language (Bias Source)}}} \\ 
        Chinese & \textbf{16.962} & \textbf{5.320} & \textbf{\textcolor{blue}{- 11.642}} \\ 
        \midrule
        \multicolumn{4}{l}{\textit{\textbf{Other Languages (Multilingual Targets)}}} \\ 
        Arabic  & 0.000 & 0.321 & + 0.321 \\ 
        English & 0.001 & 1.313 & + 1.312 \\ 
        Spanish & 0.003 & 0.736 & + 0.732 \\ 
        Hindi   & 0.001 & 0.802 & + 0.801 \\ 
        Russian & 0.102 & 1.122 & + 1.020 \\ 
        French  & 0.001 & 0.662 & + 0.661 \\ 
        Italian & 0.067 & 0.972 & + 0.906 \\ 
        Japanese& 0.003 & 0.811 & + 0.808 \\ 
        Portuguese & 0.216 & 1.112 & + 0.897 \\ 
        \midrule
        \textbf{Non-Zh Total} & \textbf{0.394} & \textbf{7.852} & \textbf{\textcolor{red}{+ 7.458}} \\ 
        \bottomrule
        \end{tabular}%
    }
    \caption{Ground-truth removal reveals same-language preference. Avg. retrieved language counts for Chinese queries ($k{=}20$, 900 queries).}
    \label{tab:language_bias_analysis}
\end{table}

\newcommand{\numrow}[1]{\addlinespace[1.5pt]#1\addlinespace[1.5pt]}
\newcommand{\best}[1]{\cellcolor{black!14}{\textbf{#1}}}
\newcommand{\lightgray}[1]{\cellcolor{black!8}{#1}}

\begin{table*}[t]
    \centering
    \small
    \renewcommand{\arraystretch}{0.8}
    \resizebox{\linewidth}{!}{%
        \begin{tabular}{lcc cc cc cc}
        \toprule
        \multirow{2}{*}{\textbf{Language}} &
        \multicolumn{2}{c}{\textbf{multilingual-e5-large}} &
        \multicolumn{2}{c}{\textbf{All-but-the-Top}} &
        \multicolumn{2}{c}{\textbf{SAE Reconstructed}} &
        \multicolumn{2}{c}{\textbf{LangSAE Editing}} \\
        \cmidrule(lr){2-3}\cmidrule(lr){4-5}\cmidrule(lr){6-7}\cmidrule(lr){8-9}
        & \textbf{nDCG@20} & \textbf{Recall@20}
        & \textbf{nDCG@20} & \textbf{Recall@20}
        & \textbf{nDCG@20} & \textbf{Recall@20}
        & \textbf{nDCG@20} & \textbf{Recall@20} \\
        \midrule

        \multicolumn{9}{c}{\textbf{Belebele}} \\
        \midrule
        \numrow{Arabic & 0.4853 & 0.4750 & 0.4194 & 0.3909 & \lightgray{0.4930} & \lightgray{0.4844} & \best{0.6810} & \best{0.6719} \\}
        \numrow{English & 0.7322 & 0.7246 & 0.7087 & 0.7028 & \lightgray{0.7370} & \lightgray{0.7298} & \best{0.7635} & \best{0.7461} \\}
        \numrow{Spanish & 0.6857 & 0.6600 & 0.6692 & 0.6458 & \lightgray{0.6888} & \lightgray{0.6633} & \best{0.7500} & \best{0.7288} \\}
        \numrow{Hindi & 0.3836 & 0.3329 & 0.3727 & 0.3209 & \lightgray{0.3884} & \lightgray{0.3393} & \best{0.4483} & \best{0.4103} \\}
        \numrow{Russian & 0.2738 & 0.1794 & 0.2631 & 0.1674 & \lightgray{0.2749} & \lightgray{0.1804} & \best{0.2766} & \best{0.1960} \\}
        \numrow{Chinese & 0.3397 & 0.2649 & \lightgray{0.4175} & \lightgray{0.3728} & 0.3461 & 0.2731 & \best{0.6947} & \best{0.6821} \\}
        \numrow{French & 0.6847 & 0.6580 & 0.6842 & 0.6569 & \lightgray{0.6918} & \lightgray{0.6656} & \best{0.7304} & \best{0.7099} \\}
        \numrow{Italian & 0.6522 & 0.6227 & 0.6416 & 0.6134 & \lightgray{0.6603} & \lightgray{0.6308} & \best{0.7485} & \best{0.7276} \\}
        \numrow{Japanese & 0.5116 & 0.4769 & \lightgray{0.5503} & \lightgray{0.5344} & 0.5171 & 0.4836 & \best{0.7119} & \best{0.7008} \\}
        \numrow{Portuguese & 0.6107 & 0.5634 & \lightgray{0.6642} & \lightgray{0.6229} & 0.6165 & 0.5712 & \best{0.7292} & \best{0.7068} \\}
        \midrule
        \numrow{\textit{Macro Average} & 0.5359 & 0.4958 & 0.5391 & \lightgray{0.5028} & \lightgray{0.5414} & 0.5022 & \best{0.6534} & \best{0.6280} \\}

        \midrule
        \addlinespace[4pt]
        \multicolumn{9}{c}{\textbf{XQuAD}} \\
        \midrule
        \numrow{Arabic & 0.6752 & 0.7557 & 0.6361 & 0.6975 & \lightgray{0.6809} & \lightgray{0.7632} & \best{0.8362} & \best{0.8972} \\}
        \numrow{English & 0.8504 & 0.9147 & 0.8210 & 0.8829 & \lightgray{0.8555} & \lightgray{0.9199} & \best{0.8751} & \best{0.9216} \\}
        \numrow{Spanish & 0.7838 & 0.8664 & \lightgray{0.7991} & 0.8709 & 0.7884 & \lightgray{0.8731} & \best{0.8672} & \best{0.9198} \\}
        \numrow{Hindi & 0.7015 & 0.7751 & \lightgray{0.7142} & 0.7777 & 0.7093 & \lightgray{0.7840} & \best{0.8443} & \best{0.8999} \\}
        \numrow{Russian & 0.7973 & 0.8908 & 0.7414 & 0.8284 & \lightgray{0.8015} & \lightgray{0.8936} & \best{0.8956} & \best{0.9457} \\}
        \numrow{Chinese & 0.4765 & 0.4831 & \lightgray{0.5854} & \lightgray{0.6392} & 0.4819 & 0.4905 & \best{0.8496} & \best{0.9080} \\}
        \midrule
        \numrow{\textit{Macro Average} & 0.7141 & 0.7810 & 0.7162 & 0.7828 & \lightgray{0.7196} & \lightgray{0.7874} & \best{0.8613} & \best{0.9154} \\}
        \bottomrule
        \end{tabular}%
    }
    \caption{MLIR performance on Belebele and XQuAD, reported by query language. We compare the base encoder, a global All-but-the-Top post-processing baseline, SAE reconstruction, and LangSAE Editing. Dark shading indicates the best result, light shading indicates the second best, computed per row and metric.}
    \label{tab:combined_results}
\end{table*}

To isolate same-language preference from semantic relevance, we measure the \emph{language distribution of retrieved distractors} \citep{yang2021simple,yang-etal-2025-language-bias-ir}. Standard retrieval metrics can obscure language bias in our setting because multiple aligned ground-truth documents exist across languages, and a system may retrieve some ground-truth items while still allocating many remaining ranks to same-language non-relevant passages. To focus on this issue, we remove aligned ground-truth documents from the \emph{retrieved list} before computing language counts.

Concretely, for each query we first retrieve the top-$20$ documents from the full multilingual pool. We then remove all aligned ground-truth documents for that query (across all included languages for that benchmark) from the retrieved list, and compute the number of remaining retrieved documents per language. After this removal, the retained documents are non-relevant under the benchmark labels, so their language distribution reflects language preference among distractors rather than the need to surface labeled answers. Because some of the original top-$20$ entries can be ground-truth documents that are removed from this analysis, the per-language counts in Table~\ref{tab:language_bias_analysis} are not expected to sum to $20$. The difference to $20$ equals the average number of ground-truth documents retrieved in the top-$20$ that were excluded from the distractor-only accounting.

We focus on Chinese queries on Belebele (900 queries). Table~\ref{tab:language_bias_analysis} shows clear evidence of same-language crowding among distractors. Under the baseline, Chinese accounts for 16.962 distractors on average, while all non-Chinese languages together account for only 0.394 distractors (17.356 distractors total). After applying \textsc{LangSAE}, the average number of Chinese distractors drops sharply to 5.320, while non-Chinese distractors increase to 7.852 (13.172 distractors total). In proportional terms, Chinese distractors drop from 97.7\% of distractors (16.962/17.356) to 40.4\% (5.320/13.172), while non-Chinese distractors rise from 2.3\% to 59.6\%. Since Belebele is parallel and the candidate pool is balanced across languages by construction, this shift is not explained by pool-size imbalance.

The gap to $20$ also increases substantially, from $20-17.356=2.644$ in the baseline to $20-13.172=6.828$ under \textsc{LangSAE}. This indicates that \textsc{LangSAE} retrieves more aligned ground-truth items within the top-$20$ while simultaneously reducing same-language crowding among the remaining non-relevant candidates. Together, these results provide direct diagnostic evidence that \textsc{LangSAE Editing} mitigates same-language preference in mixed-language retrieval pools. Qualitative retrieval examples are provided in Appendix~\ref{sec:qual_examples}.

\subsection{MLIR Retrieval Performance}
\label{sec:main_results}

Table~\ref{tab:combined_results} summarizes retrieval quality by query language on Belebele and XQuAD under our mixed-language MLIR setting, where every query retrieves from the same multilingual pool. In addition to the base encoder (\texttt{multilingual-e5-large}), we include two post-hoc baselines to separate generic embedding-space post-processing from targeted language-identity suppression: All-but-the-Top \citep{mu2018allbutthetop}, a global anisotropy-reduction transform that removes dominant principal components from the embedding space, and SAE Reconstructed, which passes embeddings through \textsc{LangSAE} without masking to isolate the effect of sparse autoencoding from feature suppression. We also observe consistent improvements when applying \textsc{LangSAE Editing} to a different multilingual embedding model (\texttt{jinaai/jina-embeddings-v3}), with results reported in Appendix~\ref{sec:appendix_jina_v3}.

Overall, \textsc{LangSAE Editing} substantially outperforms both global post-processing and reconstruction-only controls. All-but-the-Top yields small or mixed changes across languages, which is expected because it applies a single language-agnostic linear projection and does not directly target language identity. Similarly, SAE reconstruction alone provides only marginal gains, indicating that improvements are not driven by generic reconstruction effects. In contrast, masking language-associated latent units produces consistent and often large gains, supporting the claim that retrieval improvements are driven by targeted suppression of language-identity features.

First, the gains are broad rather than isolated. Improvements appear across most query languages, indicating that the method is not merely fixing a small set of pathological cases. This supports the central claim that language identity acts as a systematic shortcut in similarity search: when language-associated signal inflates same-language similarity, it affects the ordering of many competitive candidates, not only a few outliers.

Second, gains concentrate in languages that are most separable by surface form. Languages with scripts or tokenization regimes that differ sharply from the Latin-script group tend to benefit the most. This is consistent with the mechanism we target. If the encoder embeds script and orthographic cues as easily recoverable language features, then the embedding space will naturally partition by language, and nearest-neighbor retrieval will spend much of its top-$k$ capacity within the query-language region. By suppressing the latent units that behave like language identifiers, \textsc{LangSAE Editing} reduces this partitioning pressure and makes ranking depend more on shared semantic structure. The t-SNE projections in Section~\ref{sec:tsne_language_geometry} qualitatively support this interpretation.

Finally, the improvements align with our bias-focused diagnostic in Section~\ref{sec:bias_evidence}. After removing aligned ground-truth documents from the retrieved lists, the remaining distractors become less dominated by the query language, indicating that editing reduces same-language crowding among non-relevant candidates.

\paragraph{Runtime and deployment efficiency.}
Table~\ref{tab:runtime} compares the cost of post-hoc vector editing against re-running the base encoder. On 100{,}000 samples, \textsc{LangSAE Editing} takes 0.0445\,ms per sample, while the base encoder takes 0.8206\,ms, making editing \(\approx 18.4\times\) cheaper than base encoding for corpus-wide updates and only a small overhead when applied after query encoding. Masking adds 0.0131\,ms per sample over SAE reconstruction without masking. These results support our deployment claim that language-identity mitigation can be applied to stored embeddings offline and to queries at runtime with negligible compute compared to encoder tuning or corpus-wide re-encoding.

\begin{table}[t]
\centering
\setlength{\tabcolsep}{4pt}
\renewcommand{\arraystretch}{1.15}
\resizebox{\linewidth}{!}{
\begin{tabular}{lrrr}
\toprule
\textbf{Method} & \textbf{Total (100k)} & \textbf{ms / sample} & \textbf{samples / s} \\
\midrule
SAE Reconstructed & 3.1358\,s & 0.0314 & 31{,}889.37 \\
\textsc{LangSAE Editing} & 4.4516\,s & 0.0445 & 22{,}463.88 \\
multilingual-e5-large & 82.0601\,s & 0.8206 & 1{,}218.62 \\
\bottomrule
\end{tabular}
}
\caption{Runtime of post-hoc embedding transformation vs.\ base encoding, measured over 100{,}000 samples. Timings measure GPU forward-pass compute only and exclude tokenization, disk IO, and ANN search, which are identical across methods.}
\label{tab:runtime}
\end{table}

\begin{figure*}[t]
\centering
    \includegraphics[width=\linewidth]{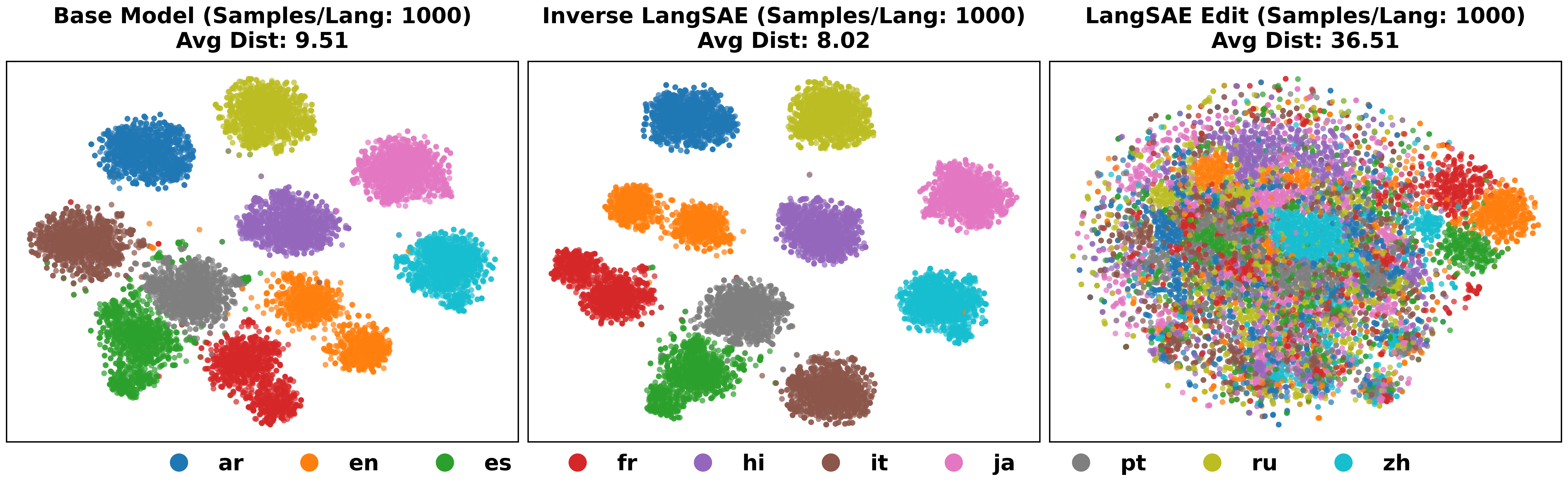}
    \caption{t-SNE projections of pooled embeddings (multilingual-e5-large, 1000 samples per language).
    \textbf{Left:} Base encoder embeddings.
    \textbf{Middle:} \textbf{Inverse mask} embeddings reconstructed using \emph{only} the language-associated units.
    \textbf{Right:} \textsc{LangSAE Edited} embeddings after suppressing the language-associated units.}
    \label{fig:tsne_language_clusters}
\end{figure*}

\subsection{Visualizing Language Identity Isolation and Removal in Embedding Space}
\label{sec:tsne_language_geometry}

Figure~\ref{fig:tsne_language_clusters} visualizes pooled embeddings from three representations using t-SNE (1000 samples per language): the base encoder space (left), an \emph{inverse-mask} reconstruction that retains only the language-associated units (middle), and the \textsc{LangSAE Edited} space after suppressing those units (right). Each panel is produced by an independent t-SNE fit. We therefore interpret the plots qualitatively in terms of separation, overlap, and local neighborhood composition, rather than absolute distances or global geometry. Note that as with any 2D projection, t-SNE can distort distances and is sensitive to hyperparameters and random seeds, but it is useful for revealing dominant clustering structure.

\paragraph{Base embeddings exhibit strong language partitioning.}
In the base encoder space (left), points form visibly language-separated regions with limited overlap. This qualitative partitioning suggests that language identity is a salient organizing factor in the pooled embedding space. In a shared multilingual retrieval pool, such separation provides an intuitive mechanism for same-language crowding: if neighborhoods are predominantly monolingual, nearest-neighbor search can preferentially traverse within-language regions, making cross-language evidence less competitive even when it is semantically relevant.

\paragraph{Inverse masking isolates a language-identifying component.}
The inverse-mask visualization (middle) reconstructs embeddings using only the latent units identified as language-associated, while zeroing all other units. In this view, language separation remains pronounced and often appears sharper than in the base space. Qualitatively, this indicates that the selected latent units capture a concentrated component that is strongly predictive of language identity, and that this component alone is sufficient to recover clear language grouping in the embedding geometry.

\paragraph{\textsc{LangSAE Editing} reduces language-driven structure.}
After suppressing the language-associated units (right), the prominent language partitioning weakens and points from different languages interleave more substantially. While t-SNE does not preserve global distances, the visible increase in cross-language overlap in local neighborhoods is consistent with the intended effect of \textsc{LangSAE Editing}: reducing language identity as a shortcut signal so that similarity neighborhoods are less dominated by language membership and can be shaped more by semantic alignment.

Overall, Figure~\ref{fig:tsne_language_clusters} provides a qualitative geometric view of the mechanism targeted by \textsc{LangSAE}. The inverse-mask panel suggests that our activation-frequency selection isolates a language-identifying component, and the edited panel shows that suppressing this component reduces language-driven clustering, consistent with the reduced same-language crowding diagnostics (Section~\ref{sec:bias_evidence}) and improved MLIR performance (Table~\ref{tab:combined_results}).

\section{Conclusion}

We studied language bias in multilingual dense retrieval, where language identity encoded in embeddings can inflate similarity for same-language pairs and crowd out relevant evidence in other languages within a shared multilingual pool. We proposed \textsc{LangSAE}, an overcomplete sparse autoencoder trained on pooled embeddings that identifies language-associated latent units via cross-language activation statistics. At inference time, \textsc{LangSAE Editing} suppresses these units and reconstructs edited embeddings in the original dimensionality, enabling retrofitting of existing vector databases without retraining the base encoder or re-encoding raw text. Experiments on mixed-language retrieval pools constructed from Belebele and XQuAD show consistent improvements in nDCG@20 and Recall@20 across languages, with diagnostic evidence that editing reduces same-language crowding among retrieved distractors. In addition to improving ranking quality, our results support a mechanistic view in which language identity occupies a small, controllable subset of sparse features that can be edited without broadly disrupting retrieval-relevant structure. Because the transformation is lightweight and vector-only, it can be applied both offline to update stored embeddings and at runtime as a small post-processing step on queries. These results indicate that language identity is a concentrated and editable factor in the representation space, and that targeted post-hoc suppression can improve MLIR in practical deployments where evidence may appear in any language.

\section*{Limitations}
Our experiments primarily evaluate \textsc{LangSAE Editing} on two parallel multilingual QA benchmarks repurposed for mixed-language retrieval (Belebele and XQuAD). While this setting provides controlled alignment across languages and enables clean diagnostics of same-language crowding, it may not capture all properties of real-world multilingual corpora, such as domain shifts, uneven language distributions, or partially overlapping relevance across languages. The method assumes access to a language label for each embedding, which is available in our benchmarks but may require metadata or language identification in deployed systems. Finally, masking behavior is controlled by an activation-frequency threshold, and while we provide a sensitivity analysis, performance can degrade if suppression becomes too aggressive. Beyond language identity, embeddings can encode other correlated nuisance factors (e.g., script, domain, formatting), and suppressing language-associated features alone may not address all sources of bias or retrieval failures.

\section*{Ethics Statement}
Our experiments use publicly available datasets and standard evaluation protocols. The underlying pretrained encoders may have been trained on large-scale web data that can contain biases, copyrighted material, or personal information beyond our control, and our work does not make claims about full pretraining data provenance. \textsc{LangSAE Editing} is intended to reduce same-language preference in multilingual retrieval, which can improve access to relevant information across languages, but it also changes the language and source distribution of retrieved results. In particular, increased cross-language retrieval may surface content that users cannot readily interpret or verify without translation, and downstream systems should consider whether to provide translation, provenance, or filtering to support safe use. Because our method is a post-hoc embedding transformation that can be applied at scale, practitioners should evaluate per-language behavior, monitor for unintended shifts in retrieval quality or exposure, and be transparent about the transformation when used in user-facing systems.

\bibliography{anthology}

\appendix

\section{Training and Implementation Details}
\label{sec:appendix_training_details}

The following details specify the \textsc{LangSAE} training configuration and the statistics used to compute activation-frequency features for language identification and masking.

\subsection{Base encoder and pooled embeddings}
\label{sec:appendix_base_encoder}

We use \texttt{intfloat/multilingual-e5-large} as the frozen base encoder. For each input text segment, token representations are mean-pooled to obtain a single embedding $\mathbf{e}\in\mathbb{R}^{d}$ with $d=1024$. \textsc{LangSAE} is trained on raw pooled embeddings (no $\ell_2$ normalization during training). At inference, reconstructed or edited embeddings are $\ell_2$-normalized before cosine similarity scoring.

\subsection{Training data construction}
\label{sec:appendix_training_data}

\paragraph{Sources and languages.}
The training corpus is constructed from mMARCO and MIRACL, restricted to 10 languages: Arabic, Chinese, English, French, Hindi, Italian, Japanese, Portuguese, Russian, and Spanish.

\paragraph{Tokenizer and segment length.}
All lengths are computed using the base encoder tokenizer. Examples with tokenized length below 250 tokens are discarded.

\paragraph{Length-based splitting.}
The goal is to expose \textsc{LangSAE} to language-identity patterns in pooled embeddings, so a length-based segmentation scheme is used. For an example with tokenized length $L$:
\begin{itemize}
    \item If $250 \le L \le 500$, keep it as a single segment.
    \item If $500 < L \le 1000$, take the first 500 tokens and split into two non-overlapping 250-token segments. Any remaining suffix shorter than 250 tokens is discarded.
    \item If $L > 1000$, partition into consecutive non-overlapping 500-token segments. Any remaining suffix shorter than 250 tokens is discarded.
\end{itemize}
All retained segments therefore fall within 250--500 tokens, and long examples yield multiple segments.

\paragraph{Balancing across languages.}
After chunking, segment counts differ across languages due to corpus variation. Each language is downsampled to match the smallest per-language segment count within each split, yielding balanced training and validation sets.

\paragraph{Train and validation sizes.}
After filtering, chunking, and balancing, the dataset contains \textbf{95{,}744{,}230} training segments and \textbf{23{,}936{,}060} validation segments. Training uses \textbf{1 epoch} over the training set.

\begin{table}[t]
\centering
\setlength{\tabcolsep}{6pt}
\renewcommand{\arraystretch}{1.1}
\resizebox{\linewidth}{!}{%
\begin{tabular}{ll}
\toprule
\multicolumn{2}{c}{\textbf{\textsc{LangSAE} checkpoint configuration}} \\
\midrule
Base encoder & \texttt{multilingual-e5-large} \\
Training corpora & mMARCO + MIRACL \\
Embedding dimension $d$ & 1024 \\
Expansion factor $m/d$ & 256 \\
Dictionary size $m$ & 262{,}144 \\
Sparsity (top-$k$) & 4{,}096 \\
Learning rate & $5\times 10^{-4}$ \\
Auxiliary loss coefficient & $1\times 10^{-1}$ \\
Auxiliary usage target & $2\times 10^{-2}$ \\
Epochs & 1 \\
\midrule
\multicolumn{2}{c}{\textbf{Training and validation summary}} \\
\midrule
Aux loss (train / val) & 0.9594 / 0.9719 \\
Dead features (\%) (train / val) & 0 / 0 \\
FVU (train / val) & 0.002116 / 0.002109 \\
$\ell_0$ active features (train / val) & 3454.35 / 3457.18 \\
Total loss (train / val) & $5.60\times 10^{-7}$ / $5.58\times 10^{-7}$ \\
MSE loss (train / val) & $5.60\times 10^{-7}$ / $5.58\times 10^{-7}$ \\
\bottomrule
\end{tabular}%
}
\caption{Top: \textsc{LangSAE} checkpoint configuration used in main experiments. Bottom: training and validation summary for the same run.}
\label{tab:langsae_config_and_summary}
\end{table}

\subsection{Probe set for activation-frequency statistics}
\label{sec:appendix_probe_set}

To compute activation frequencies $p_{i,\ell}$ for language-feature identification (Section~\ref{sec:lang_feature_identification}), we use the same validation split described above, grouped by language. Specifically, we embed each validation segment with the frozen encoder, compute its sparse code $\tilde{\mathbf{z}}$, and estimate
$p_{i,\ell}=\mathbb{E}_{t\sim\mathcal{D}_\ell}[\mathbb{I}(\tilde{z}_i(t)>0)]$
using validation segments $t$ in language $\ell$.

\subsection{Auxiliary feature-usage loss}
\label{sec:appendix_aux_loss}

In addition to reconstruction loss (Eq.~\ref{eq:recon_loss}), we employ an auxiliary feature-usage encouragement objective to mitigate the dead-feature problem in top-$k$ sparse autoencoders. Intuitively, this loss penalizes latent units whose estimated activation frequency falls below a user-defined minimum usage target, encouraging the model to utilize the full dictionary capacity instead of collapsing onto a small subset of features.

Concretely, we define a per-unit \emph{activation deficit} as the non-negative difference between a target activation fraction and the unit's estimated activation frequency. Activation frequencies are estimated differentiably (via a high-temperature sigmoid surrogate), and the auxiliary loss is computed as the mean squared activation deficit across units. The auxiliary coefficient and target used in the main checkpoint are reported in Table~\ref{tab:langsae_config_and_summary}.

\subsection{Optimization, precision, and hardware}
\label{sec:appendix_optimization}

We optimize \textsc{LangSAE} with Adam. Training uses mixed precision (fp16) on CUDA via \texttt{torch.cuda.amp.autocast} and \texttt{GradScaler} for numerical stability. Training was performed on 8 NVIDIA RTX A6000 GPUs. For the expansion-factor-256 configuration used in our main experiments, training took approximately 6 hours.

\subsection{\textsc{LangSAE} training summary}
\label{sec:appendix_sae_config_summary}

Table~\ref{tab:langsae_config_and_summary} reports the configuration and logged statistics for the \textsc{LangSAE} checkpoint used in the main experiments. The overcomplete dictionary size is determined by the expansion factor ($m/d$), and sparsity is enforced by top-$k$ selection on ReLU activations (Section~\ref{sec:langsae_training}). Reported training statistics include the auxiliary loss used to encourage feature usage, the dead-feature rate, and reconstruction quality measured by fraction of variance unexplained (FVU). The $\ell_0$ statistic corresponds to the number of non-zero latent activations per example after top-$k$ sparsification.

\section{Evaluation Benchmarks}
\label{sec:appendix_benchmarks}

We evaluate multilingual retrieval in mixed-language pools using multilingual question answering (QA) datasets with parallel constructions, repurposed as retrieval tasks. These datasets provide aligned query and document instances across languages, enabling controlled cross-language evaluation without relying on heuristic relevance transfer. Since the original task is extractive QA, the associated passages serve as precise gold evidence for retrieval: for each question, the passage paired with that question is treated as relevant, and parallel variants of that passage across languages define aligned relevant evidence under our multilingual pool setting. This evaluation paradigm is widely used in recent work to assess multilingual and cross-lingual retrieval behavior using QA resources with parallel structure.

\paragraph{Belebele.}
Belebele~\citep{bandarkar-etal-2024-belebele} is a professionally translated multilingual QA dataset designed to support high-quality multilingual evaluation across a broad set of languages. Translations were produced by native speakers proficient in English, aiming to preserve both contextual meaning and language-specific nuances. In our retrieval formulation, we treat each passage as a document and each question as a query. Because passages are parallel across languages via passage identifiers, we can construct a multilingual candidate pool by taking the union of passages across languages and define a multi-relevant set for each query consisting of all aligned passages across the included languages. This parallel structure makes Belebele well-suited for diagnosing same-language preference and measuring whether a retriever surfaces semantically aligned evidence across languages in a shared multilingual pool.

\begin{table*}[t]
    \centering
    \tiny
    \renewcommand{\arraystretch}{0.8}
    \resizebox{\linewidth}{!}{%
        \begin{tabular}{lcc cc cc}
        \toprule
        \multirow{2}{*}{\textbf{Language}} &
        \multicolumn{2}{c}{\textbf{jina-embeddings-v3}} &
        \multicolumn{2}{c}{\textbf{SAE Reconstructed}} &
        \multicolumn{2}{c}{\textbf{LangSAE Editing}} \\
        \cmidrule(lr){2-3}\cmidrule(lr){4-5}\cmidrule(lr){6-7}
        & \textbf{nDCG@20} & \textbf{Recall@20}
        & \textbf{nDCG@20} & \textbf{Recall@20}
        & \textbf{nDCG@20} & \textbf{Recall@20} \\
        \midrule

        \multicolumn{7}{c}{\textbf{Belebele}} \\
        \midrule
        \numrow{Arabic & \best{0.5504} & \lightgray{0.5574} & 0.5450 & 0.5530 & \lightgray{0.5474} & \best{0.5586} \\}
        \numrow{English & 0.5806 & \best{0.5876} & \lightgray{0.5836} & \lightgray{0.5874} & \best{0.5839} & 0.5873 \\}
        \numrow{Spanish & \lightgray{0.6649} & \lightgray{0.6739} & 0.6624 & 0.6690 & \best{0.6653} & \best{0.6740} \\}
        \numrow{Hindi & 0.3752 & 0.3656 & \lightgray{0.3782} & \lightgray{0.3666} & \best{0.3802} & \best{0.3683} \\}
        \numrow{Russian & 0.2071 & 0.1760 & \lightgray{0.2103} & \lightgray{0.1769} & \best{0.2111} & \best{0.1774} \\}
        \numrow{Chinese & 0.5529 & 0.5430 & \lightgray{0.5611} & \lightgray{0.5538} & \best{0.5648} & \best{0.5617} \\}
        \numrow{French & \lightgray{0.6112} & \best{0.6186} & 0.6108 & 0.6159 & \best{0.6134} & \best{0.6186} \\}
        \numrow{Italian & \lightgray{0.6488} & \lightgray{0.6529} & 0.6471 & 0.6510 & \best{0.6520} & \best{0.6568} \\}
        \numrow{Japanese & \lightgray{0.5786} & 0.5757 & 0.5767 & \lightgray{0.5768} & \best{0.5809} & \best{0.5828} \\}
        \numrow{Portuguese & \lightgray{0.6318} & \best{0.6388} & 0.6304 & 0.6354 & \best{0.6333} & \best{0.6388} \\}
        \midrule
        \numrow{\textit{Macro Average} & 0.5401 & \lightgray{0.5390} & \lightgray{0.5405} & 0.5386 & \best{0.5432} & \best{0.5423} \\}

        \midrule
        \addlinespace[4pt]
        \multicolumn{7}{c}{\textbf{XQuAD}} \\
        \midrule
        \numrow{Arabic & 0.7156 & 0.7894 & \lightgray{0.7272} & \lightgray{0.7976} & \best{0.7302} & \best{0.8048} \\}
        \numrow{English & 0.7461 & 0.8169 & \best{0.7558} & \best{0.8234} & \lightgray{0.7505} & \lightgray{0.8221} \\}
        \numrow{Spanish & 0.7927 & 0.8578 & \lightgray{0.7980} & \lightgray{0.8595} & \best{0.8016} & \best{0.8651} \\}
        \numrow{Hindi & 0.7334 & 0.8003 & \lightgray{0.7480} & \lightgray{0.8113} & \best{0.7509} & \best{0.8181} \\}
        \numrow{Russian & 0.7616 & 0.8293 & \lightgray{0.7725} & \lightgray{0.8373} & \best{0.7742} & \best{0.8413} \\}
        \numrow{Chinese & 0.6950 & 0.7660 & \lightgray{0.7171} & \lightgray{0.7849} & \best{0.7253} & \best{0.7952} \\}
        \midrule
        \numrow{\textit{Macro Average} & 0.7408 & 0.8101 & \lightgray{0.7530} & \lightgray{0.8191} & \best{0.7556} & \best{0.8247} \\}
        \bottomrule
        \end{tabular}%
    }
    \caption{MLIR performance on Belebele and XQuAD using jina-embeddings-v3, reported by query language. Dark shading indicates the best result, light shading indicates the second best, computed per row and metric.}
    \label{tab:combined_results_jina}
\end{table*}

\paragraph{XQuAD.}
XQuAD~\citep{artetxe-etal-2020-cross} is a multilingual QA benchmark derived from SQuAD 1.1~\citep{rajpurkar-etal-2016-squad}. It provides translations of question-answer pairs and context paragraphs into multiple languages, yielding fully parallel examples across languages. In our retrieval formulation, each translated context paragraph is treated as a document and each translated question is treated as a query. The strict one-to-one alignment across languages enables constructing multilingual pools and defining aligned relevant document sets analogously to Belebele. This makes XQuAD a useful benchmark for evaluating the stability of embedding-based retrieval under linguistic variation and for measuring how language-identity signal affects similarity search in multilingual settings.

\paragraph{Why parallel QA datasets.}
We require a benchmark construction where, for every query, there is guaranteed relevant evidence in multiple languages within a single shared candidate pool. This is essential for diagnosing \emph{same-language crowding} cleanly: we need cross-language relevant documents to exist by construction so that a retrieval failure under a mixed-language pool can be attributed to language-identity effects rather than missing cross-language relevance labels or incomplete cross-language annotation. Parallel QA datasets provide this property through their one-to-one alignment across languages, allowing us to define multi-relevance sets that include all aligned passages for each query and to evaluate whether retrieval surfaces semantically matching evidence beyond the query language.

We considered a broader range of multilingual retrieval datasets, but many multilingual IR benchmarks are designed primarily for \emph{monolingual} retrieval within each language and therefore do not provide fully parallel, one-to-one aligned query--document pairs across languages. In particular, Mr.~TyDi~\citep{zhang-etal-2021-mr} and MIRACL~\citep{zhang-etal-2023-miracl} contain language-specific query sets and relevance judgments over language-specific corpora, rather than a shared pool with guaranteed cross-language aligned relevant documents for every query. This makes it non-trivial to construct controlled multi-relevance sets in mixed-language pools without introducing additional cross-language alignment machinery (e.g., entity or document linking). Because our goal is to isolate and measure same-language preference in a setting where cross-language relevant evidence is present by construction, we focus on Belebele and XQuAD as our primary evaluation datasets.

\section{Additional Encoder Results: \texttt{jinaai/jina-embeddings-v3}}
\label{sec:appendix_jina_v3}

To evaluate whether \textsc{LangSAE Editing} generalizes beyond \texttt{multilingual-e5-large}, we repeat the mixed-language MLIR protocol from Section~\ref{sec:experiments} using \texttt{jinaai/jina-embeddings-v3} as the frozen base encoder. We train a separate \textsc{LangSAE} on pooled embeddings produced by this encoder, then apply the same inference-time editing procedure to both query and document vectors. Unless stated otherwise, we use an expansion factor of 128 with top-$k{=}2048$, learning rate $3\times 10^{-4}$, auxiliary coefficient $10^{-1}$, and usage target $2\times 10^{-2}$. At inference we use the Unique+Overlapping masking strategy with $\tau{=}0.999$.

Table~\ref{tab:combined_results_jina} reports nDCG@20 and Recall@20 by query language on Belebele and XQuAD. Overall, the post-hoc transformation yields consistent, if smaller, improvements compared to the base encoder, indicating that the language-identity signal exploited by similarity search is not specific to a single encoder family and that sparse feature suppression can provide benefits across encoder architectures.

\section{Sensitivity to Activation-Frequency Threshold \texorpdfstring{$\tau$}{tau}}
\label{sec:ablation_tau}

\begin{table}[h]
    \centering
    \resizebox{\linewidth}{!}{%
        \begin{tabular}{l|cc|cc}
        \toprule
        \multicolumn{5}{c}{\textbf{multilingual-e5-large}} \\
        \midrule
         & \multicolumn{2}{c|}{\textbf{Belebele (Macro Avg)}} & \multicolumn{2}{c}{\textbf{XQuAD (Macro Avg)}} \\
        \textbf{Threshold} & \textbf{nDCG@20} & \textbf{Recall@20} & \textbf{nDCG@20} & \textbf{Recall@20} \\
        \midrule
        1.000 & 0.5974 & 0.5717 & 0.7876 & 0.8669 \\
        0.999 & 0.6534 & 0.6280 & 0.8613 & 0.9154 \\
        0.998 & 0.6019 & 0.5767 & 0.8258 & 0.8775 \\
        0.997 & 0.4817 & 0.4615 & 0.7369 & 0.7843 \\
        0.996 & 0.2768 & 0.2708 & 0.5130 & 0.5640 \\
        0.995 & 0.0874 & 0.0939 & 0.2043 & 0.2505 \\
        0.990 & 0.0455 & 0.0481 & 0.1495 & 0.1742 \\
        \bottomrule
        \end{tabular}%
    }
    \caption{Sensitivity to activation-frequency threshold $\tau$ (absolute macro-average).}
    \label{tab:tau_ablation}
\end{table}

The activation-frequency threshold $\tau$ controls how conservatively \textsc{LangSAE Editing} selects latent units for suppression. Using per-language activation frequencies $p_{i,\ell}$ (Eq.~\ref{eq:activation_frequency}), we define frequent sets as $\mathcal{F}_\ell(\tau)=\{i \mid p_{i,\ell}\ge \tau\}$, derive language-unique and overlapping sets $\mathcal{U}_\ell(\tau)$ and $\mathcal{O}(\tau)$ as in Section~\ref{sec:lang_feature_identification}, and apply the same masking strategy used elsewhere:
\begin{equation}
\mathcal{S}_\ell(\tau) = \mathcal{U}_\ell(\tau)\cup \mathcal{O}(\tau).
\end{equation}

Table~\ref{tab:tau_ablation} reports \emph{absolute} macro-average nDCG@20 and Recall@20 under mixed-language retrieval in multilingual pools. Performance exhibits a narrow high-performing band near $\tau \approx 0.999$--$1.000$: $\tau=0.999$ achieves the strongest results on both benchmarks, and $\tau\in\{1.000,0.998\}$ remains competitive. However, once $\tau$ is relaxed further, performance degrades rapidly. By $\tau=0.997$ metrics drop substantially, and by $\tau\le 0.995$ retrieval quality collapses to very low values under our multi-relevance evaluation, indicating that masking has removed substantial retrieval-relevant structure.

\begin{figure}[t]
    \centering
    \includegraphics[width=\linewidth]{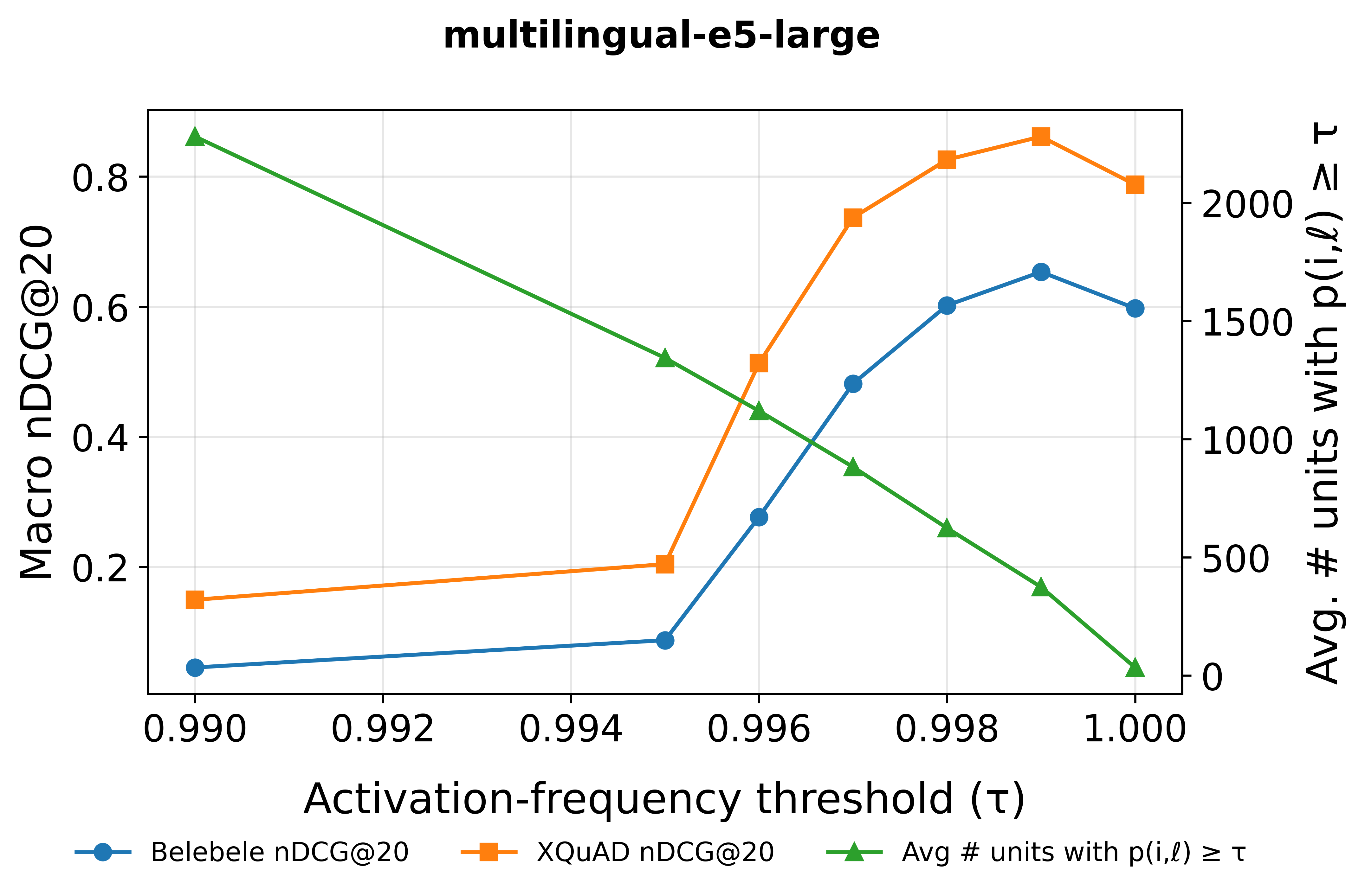}
    \caption{Sensitivity to $\tau$. Macro nDCG@20 as a function of the activation-frequency threshold $\tau$ (left axis), together with the average number of latent units per language whose activation frequency exceeds $\tau$ (right axis). As $\tau$ decreases slightly below 1.0, the set of frequently active units grows rapidly, which propagates to a much larger suppression set and can trigger over-masking.}
    \label{fig:tau_sensitivity}
\end{figure}

Figure~\ref{fig:tau_sensitivity} clarifies why small changes in $\tau$ can have outsized effects. The right axis shows that the number of latent units with $p_{i,\ell}\ge\tau$ increases sharply as $\tau$ decreases in the narrow region just below 1.0. This reflects a concentration of activation frequencies near one, which is expected in a top-$k$ sparse autoencoder where a subset of features is reused consistently across many inputs. Because $\mathcal{S}_\ell(\tau)$ is built from frequent sets (and includes both language-unique and overlapping frequent units), this rapid growth propagates into a much larger suppression set. Past a critical point, masking begins to remove not only language-associated shortcut features but also frequently used factors that support semantic similarity, which contracts similarities for both positives and competitive negatives and destroys the relative separability required for accurate ranking.

Overall, these results show that activation-frequency thresholding must be used conservatively. Values in a tight neighborhood near $\tau\approx 0.999$--$1.000$ can suppress highly consistent language-associated units while preserving most shared semantic structure, whereas modest additional relaxation triggers over-masking and severe degradation.

\section{Overlap Removal vs.\ Non-Removal}
\label{sec:ablation_overlap}

Overlap handling controls whether \textsc{LangSAE Editing} suppresses only language-unique features, or suppresses both language-unique and overlapping features that are frequent across multiple languages. The comparison uses three settings: (i) no feature suppression (Baseline), (ii) suppression of language-unique features only, and (iii) suppression of language-unique plus overlapping features. Table~\ref{tab:ablation_detailed} reports macro-average results on Belebele under the same evaluation protocol as the main experiments.

Including overlapping features in the suppression set yields the best overall performance, while masking language-unique features alone can degrade retrieval relative to the baseline. This suggests that language bias in the embedding space is not carried exclusively by language-unique units. Instead, a substantial portion of the shortcut signal appears to live in features that are frequent across multiple languages, for example shared script- or tokenization-related regularities, or multilingual corpus artifacts, that can still inflate cosine similarity and contribute to same-language crowding. Masking only language-unique units may therefore remove some retrieval-relevant variation without sufficiently attenuating the dominant cross-language shortcut features, whereas additionally suppressing overlapping features more effectively weakens language-driven similarity and improves ranking in mixed-language pools.

\section{Qualitative Ranking Examples}
\label{sec:qual_examples}

\newcolumntype{L}[1]{>{\raggedright\arraybackslash}p{#1}}
\newcolumntype{C}[1]{>{\centering\arraybackslash}p{#1}}
\newcommand{\markO}{\textcolor{blue}{\textbf{O}}\xspace}
\newcommand{\markX}{\textcolor{red}{\textbf{X}}\xspace}

To complement the aggregate metrics, we present two qualitative examples that show how \textsc{LangSAE Editing} changes the composition of the top-ranked results in mixed-language pools. In both cases, the baseline retrieval list is dominated by same-language items, and several of these high-ranked candidates are distractors that are only weakly related to the query. After editing, the ranking surfaces additional aligned evidence written in other languages, increasing cross-language coverage while preserving the ability to retrieve relevant passages in the query language. These examples align with our quantitative analysis of same-language crowding (Table~\ref{tab:language_bias_analysis}).

\begin{table}[t]
    \centering
    \small
    \resizebox{\linewidth}{!}{%
        \begin{tabular}{l|cc}
        \toprule
        \multicolumn{3}{c}{\textbf{multilingual-e5-large}} \\
        \midrule
         & \multicolumn{2}{c}{\textbf{Average}} \\
        \textbf{Removal Strategy} & \textbf{nDCG@20} & \textbf{Recall@20} \\ \midrule
        No Removal & 0.5359 & 0.4958 \\
        Unique Only & 0.5176 & 0.4696 \\
        Unique + Overlapping & \textbf{0.6534} & \textbf{0.6280} \\
        \bottomrule
        \end{tabular}%
    }
    \caption{Macro-average results on Belebele under different suppression strategies. Suppressing overlapping features in addition to language-unique features yields the strongest MLIR performance.}
    \label{tab:ablation_detailed}
\end{table}

Markers indicate whether the retrieved passage is aligned ground-truth evidence for the query (\markO) or not (\markX).

\begin{CJK}{UTF8}{gbsn}
\begin{table*}[htbp]
\centering

\renewcommand{\arraystretch}{1.20}
\setlength{\tabcolsep}{5pt}

\resizebox{\linewidth}{!}{
\begin{tabular}{C{0.06\textwidth} L{0.41\textwidth} C{0.05\textwidth} L{0.41\textwidth} C{0.05\textwidth}}
\toprule

\multicolumn{5}{L{0.98\linewidth}}{\textbf{Query (ZH):} 根据这段文字，亚马逊河的河水来自哪里？} \\
\multicolumn{5}{L{0.98\linewidth}}{\textbf{English:} Based on this text, where does the water of the Amazon River come from?} \\
\addlinespace[3pt]
\midrule

\textbf{Rank} &
\multicolumn{1}{c}{multilingual-e5-large} &
\textbf{Rel.} &
\multicolumn{1}{c}{\textsc{LangSAE}} &
\textbf{Rel.} \\
\midrule

1 &
亚马逊河是世界上第二长，也是最大的河流。它的水量是第二大河流的 8 倍以上... &
\markO &
亚马逊河是世界上第二长，也是最大的河流。它的水量是第二大河流的 8 倍以上... &
\markO \\
\addlinespace

2 &
1963 年大坝建成后，季节性洪水被控制住了，沉积物不再冲散到河流里... &
\markX &
[PT] O Amazonas é o maior rio e o segundo mais longo da Terra... &
\markO \\
\addlinespace

3 &
维京人利用俄罗斯水路到达黑海和里海。其中一些路线至今仍可通行... &
\markX &
[EN] The Amazon River is the second longest and the biggest river... &
\markO \\
\addlinespace

4 &
印度河流域文明是青铜时代的文明，位于印度西北部次大陆... &
\markX &
[FR] Le fleuve Amazone est le deuxième plus long et le plus grand... &
\markO \\
\addlinespace

5 &
联合国维和人员在 2010 年地震后抵达海地，他们因疫情蔓延而受到指责... &
\markX &
[ES] El río Amazonas es el más caudaloso y el segundo más extenso... &
\markO \\
\bottomrule
\end{tabular}
}
\caption{Qualitative retrieval example (Amazon River query). \markO indicates the passage contains the correct evidence, \markX otherwise.}
\label{tab:qual_amazon}
\end{table*}
\end{CJK}

\begin{CJK}{UTF8}{gbsn}
\begin{table*}[htbp]
\centering

\renewcommand{\arraystretch}{1.20}
\setlength{\tabcolsep}{5pt}

\resizebox{\linewidth}{!}{
\begin{tabular}{C{0.06\textwidth} L{0.41\textwidth} C{0.05\textwidth} L{0.41\textwidth} C{0.05\textwidth}}
\toprule

\multicolumn{5}{L{0.98\linewidth}}{\textbf{Query (ES):} Según el texto, ¿cuál de las siguientes opciones no se recomienda para que los atletas jóvenes disfruten más el deporte?} \\
\multicolumn{5}{L{0.98\linewidth}}{\textbf{English:} According to the text, which of the following is not recommended for young athletes to enjoy sports more?} \\
\addlinespace[3pt]
\midrule

\textbf{Rank} &
\multicolumn{1}{c}{multilingual-e5-large} &
\textbf{Rel.} &
\multicolumn{1}{c}{\textsc{LangSAE}} &
\textbf{Rel.} \\
\midrule

1 &
No es posible que las prácticas nutricionales adecuadas, por sí solas, generen un rendimiento de elite... &
\markO &
No es posible que las prácticas nutricionales adecuadas, por sí solas, generen un rendimiento de elite... &
\markO \\
\addlinespace

2 &
[PT] A nutrição adequada por si só não gera desempenhos de alta performance, mas pode afetar... &
\markO &
[PT] A nutrição adequada por si só não gera desempenhos de alta performance, mas pode afetar... &
\markO \\
\addlinespace

3 &
La carrera de distancia media es un deporte relativamente económico; no obstante... &
\markX &
La carrera de distancia media es un deporte relativamente económico; no obstante... &
\markX \\
\addlinespace

4 &
USA Gymnastics respalda la nota del Comité Olímpico de los Estados Unidos... &
\markX &
[ZH] 仅靠适当的营养实践并不足以造就出色表现，但这可以显著影响年轻运动员... &
\markO \\
\addlinespace

5 &
El ganador olímpico de la medalla de oro debía nadar en el estilo libre de 100 metros... &
\markX &
[IT] Le sole pratiche nutrizionali corrette non bastano a generare elevate prestazioni... &
\markO \\
\bottomrule
\end{tabular}
}
\caption{Qualitative retrieval example (athlete nutrition query). \markO indicates the passage contains the correct evidence, \markX otherwise.}
\label{tab:qual_athlete}
\end{table*}
\end{CJK}

\end{document}